# Research on the Laws of Multimodal Perception and Cognition from a Cross-cultural Perspective --- Taking Overseas Chinese Gardens as an Example

Ran Chen* [1], Xueqi Yao* [1], Jing Zhao[1], Shuhan Xu[1], Sirui Zhang[1], Yijun Mao[1]

**Abstract**: This study aims to explore the complex relationship between perceptual and cognitive interactions in multimodal data analysis, with a specific emphasis on spatial experience design in overseas Chinese gardens. It is found that evaluation content and images on social media can reflect individuals' concerns and sentiment responses, providing a rich data base for cognitive research that contains both sentimental and image-based cognitive information. Leveraging deep learning techniques, we analyze textual and visual data from social media, thereby unveiling the relationship between people's perceptions and sentiment cognition within the context of overseas Chinese gardens. In addition, our study introduces a multi-agent system (MAS) alongside AI agents. Each agent explores the laws of aesthetic cognition through chat scene simulation combined with web search. This study goes beyond the traditional approach of translating perceptions into sentiment scores, allowing for an extension of the research methodology in terms of directly analyzing texts and digging deeper into opinion data. This study provides new perspectives for understanding aesthetic experience and its impact on architecture and landscape design across diverse cultural contexts, which is an essential contribution to the field of cultural communication and aesthetic understanding.

1 School of Landscape Architecture, Beijing Forestry University, Beijing, 100083, China

*Ran Chen and Xueqi Yao contributed equally to this manuscript

## 1 Introduction

The relationship between perception and cognition is one of the central issues in spatial experience design. When people are in a physical space, their perceptual and cognitive activities are activated. In perception, humans receive and process information about the external environment. In cognitive activities, the human brain further processes external information and evokes pleasurable emotions. Therefore, the study of spatial experience focuses on the correlation between human perception and cognition.

In recent years, researchers have delved into the complex interactions between perception and cognitive responses. The primary focus of research in this field lies in understanding how individuals perceive spatial environments through their senses and subsequently translate these perceptions into sentimental and cognitive responses. This endeavor aims to uncover the dynamic interplay between perception and cognition. Especially in the field of the built environment, this correspondence is one of the most important criteria for judging design works. Overseas Chinese gardens, as unique examples of built environments, provide an ideal platform for research. These gardens are not only an international presentation of Chinese garden art, but also integrate local cultural characteristics, becoming an important medium for cross-cultural aesthetics and cultural dissemination. By studying the perceptual and cognitive interactions in overseas Chinese gardens, we can gain an in-depth understanding of aesthetic experiences in different cultural contexts and their impact on architecture and landscape design. In addition, the study of these gardens not only enriches the theories in the field of the built environment, but also provides new perspectives on international cultural exchanges and multicultural cohesion. Consequently, it holds significant implications for both academic and cultural communication fields.

In reviewing existing research, it is found that perception is often characterized by visual images, such as real-life photographs and images captured on the Internet, while cognition is mostly presented in the form of text, such as online reviews and questionnaire data. However, most studies have focused on unimodal studies, i.e., focusing only on text or images. Given that the interaction between perception and cognition involves multimodal information in the form of graphic and textual intersections, this calls for innovations at both the level of data acquisition and data analysis methods.

This study addresses the research problem of perception-cognition interaction in multimodal data analysis and innovatively proposes a method based on social media opinion data. By comparing the data characteristics of social media and traditional news media, we find that the evaluation content posted by users in social media and their attached images can reflect individuals' key concerns and deep impressions, while displaying strong personal emotional responses. This provides a rich data base for cognitive research that contains both emotional and image cognitive information. Furthermore, we have explored the application of emerging deep learning techniques in handling such multimodal data. Particularly, in large-scale studies concerning the relationship between text and images, deep learning methods have demonstrated their effectiveness.

In traditional aesthetic cognition research, researchers usually translate cognition into sentiment scores and analyze the relationship between numbers and figures. However, with the introduction of multi-agent systems (MAS) and the application of AI agents, the research methodology has been extended. In this study, we established a discussion community based on the MAS, a community in which individual agents conducted web searches through chatting combined with networking to explore the scientific law of aesthetic perception of overseas Chinese gardens.

The main contribution of this study is to provide a new perspective and methodology for exploring the application of social media data in multimodal perception and cognition research. We chose overseas Chinese gardens as the research object, and by analyzing the data from five major foreign opinion websites and using multimodal techniques, we explored the relationship between people's perception (through images) and emotional cognition (through text) of specific garden scenes. Through the application of deep learning techniques, we not only extend the methodology of multimodal data analysis, but also provide new data, perspectives and cognitive insights for the study of overseas Chinese gardens. this, in turn, opens up new paths for the study of cultural communication and aesthetic experience.

The rest of the paper is organized as follows. Section 2 conducts a literature review of related work. Section 3 describes our architecture and training methodology. Section 4 describes the experiments as well as the results of the tests. Section 5 presents a discussion of related studies. Section 6 elucidates the results and future related work.

## 2 related works

In the context of integrating current research trends, our objective is to employ deep learning techniques to explore the relationship between perception and cognition, using overseas Chinese gardens as an example. Our review found that although many studies have explored perception based on images(Serap Yılmaz, 2018; Zhang et al., 2022), as well as texts for cognitive research(Sourav & Wang, 2022a; Kusal et al., 2021; Shah & R, 2022a), and studies on aesthetic preferences have gradually emerged, few studies have focused on the aesthetic-cognitive patterns of overseas Chinese gardens, especially when dealing with large amounts of opinion data.

In addition, we have constructed an intelligent agent, termed an 'aesthetic-agent, specialized in processing aesthetic cognitive laws based on the latest large language modeling technology.

### 2.1 History and Current Status of Perception Research

#### 2.1.1 Overview of traditional perception research (e.g., research on the senses of sight, hearing, etc.)

Traditional perception research has focused on laboratory settings using a variety of instruments to understand how the basic human senses such as vision, hearing, and touch work and how they interact to affect our perception of the world. These studies employ a variety of methods including observational methods, experimental methods, psychophysical tests, behavioral experiments, and neuroimaging techniques (FMRI,EEG). Through these methods, researchers gain a deeper understanding of the biological and psychological mechanisms by which the senses work and develop applications and assistive technologies that can improve the quality of life for individuals with sensory impairments.

#### 2.1.2 Applications of image perception in spatial design (with special reference to urban data and streetscape perception)

Now in the field of built environment, researchers are also starting to study how the public perceives the built environment. Current research is mainly based on deep learning techniques to perceive street scenes or urban spaces.

First for street space perception, research in this area has focused on using deep learning and computer vision techniques to quantify and recognize physical and psychological features of urban streetscape images. For example, Bocco's (2021) [Bocco, 2021] research focuses on the quantification of perceived urban safety on streetscape images, while Rodan (2022) (Xu et al., 2022)explores the relationship between streetscape perception and housing prices, using computer vision and machine learning techniques for subjective and objective measurements. These studies provide new perspectives for understanding the safety, aesthetic, and economic value of urban spaces.

Secondly for the perception of other urban spaces, researchers have used machine learning and human-city-environment synergistic perception methods to assess urban spatial perception, which in turn provides new ideas for urban spatial design. For example, Haohao Ji et al. proposed a novel data intelligence framework for assessing urban spatial perception, which demonstrates the advancement of data analytics in understanding and evaluating urban environments(Ji et al., 2021). Zhao Jingzhu's (2021) research quantifies the physical and psychological perception of urban scenes through deep learning techniques, aiding urban planners in more detailed city design from residents' perspectives. (Guan et al., 2021). Yonghong Peng's (2023) study looks at machine learning and collaborative human-city-environment perception methods, and its application to case domains exemplifies the broad applicability of AI in urban perception research (Huang et al., 2023).

Taken together, these recent studies provide further demonstrate the importance and application potential of deep learning and computer vision technologies in urban space design and planning. Including the deeper integration of the technologies with everyday experiences and the enhancement of sustainability and inclusiveness in urban planning.

### 2.2 Cognitive Theory and Emotion Research

In current cognitive emotion research, text-based emotion detection has become mainstream. Researchers have employed various techniques to extract and recognize emotional information from textual data. For example, Sheetal Kusal et al. (2021) reviewed text-based emotion detection techniques, emphasizing the use of artificial intelligence in textual sentiment analysis(Kusal et al., 2021); Pradeep Kumar Shah (2022) explored the use of artificial intelligence to model feedback systems for emotion interpretation(Shah & R, 2022); while Md Sakib Ullah Sourav et al. (2022) used the Transformer model to classify emotional texts in Bengali(Sourav & Wang, 2022). With these techniques, researchers are able to identify emotional states from large amounts of textual data and apply them to market trend analysis, consumer behavior research, and spatial design(Shah & R, 2022; Kusal et al., 2021; Sourav & Wang, 2022).

In particular, in the area of the interaction between emotional cognition and spatial design, current research has focused on investigating the use of data obtained from textual analysis to optimize spatial design with the aim of better meeting the emotional and psychological needs of users. In the area of textual sentiment analysis, most of the research has focused extensively on destination image analysis, and the development of the CARE dataset by Jane A. Yu and A. Halevy (2022) has provided a new resource for emotional response detection(Yu & Halevy, 2022).

In terms of spatial design, the interaction between emotional cognition and spatial design is gradually gaining attention, with studies showing that the details of spatial design affect people's emotional experience, and people's emotional state changes their perception and experience of space (Higuera-Trujillo et al., 2021; Yu & Halevy, 2022)(Jane A. Yu, A. Halevy, 2022, Higuera, 2021). Text-based sentiment detection has provided new perspectives and tools for cognitive emotion research, and these studies have further demonstrated the potential and value of text-based sentiment analysis in a variety of application scenarios, especially in processing big data and optimizing spatial experience design(Yu & Halevy, 2022; Kusal et al., 2021; Shah & R, 2022b; Sourav & Wang, 2022b). Future research will continue to explore and improve the accuracy and potential for cross-domain applications of emotion detection techniques.To summarize, in the overview of emotional cognition research, most current studies focus on text-based emotional analysis of tourism destination images and spatial design perception analysis. There is still a research gap in how current studies combine insights from textual sentiment analysis with actual spatial experience design practice.

## 2.3 Interaction between Perception and Cognition

### 2.3.1 Relationship between perception and cognition (theoretical perspective)

In the field of cognitive psychology, the relationship between perception and cognition is seen as interactive and interconnected. Perception, as the primary means of acquiring information about the outside world, provides the underlying data for cognitive activities, while cognition-higher mental processes involving thinking and decision-making-in turn shapes and influences perception. Theoretical studies have shown that cognitive processes such as thought categorization can alter perceptual details, reflecting the fact that perception and cognition are intertwined(Brown et al., 2011; Goldstone et al., 2011) . This dynamic interaction refers to the fact that cognitive structures are enriched by perceptual experiences, and perceptual inputs are adjusted by cognitive processes.

### 2.3.2 Specific application cases of perception and cognition in spatial experience design

In studies related to the built environment, scholars primarily focus on aesthetic preferences and their influencing factors regarding the relationship between perception and cognition. Covering a wide range of environments from riverside landscapes (Zhang et al., 2022) to urban parks (Serap Yılmaz, 2018), these studies have explored how aesthetic preferences are interrelated with spatial elements, socio-demographics, and individual activities such as eating (Sakıcı, 2022). Conjoint analysis was used to gain insight into aesthetic preferences (Schirpke et al., 2019), while sociodemographic influences were examined through a study of visual landscape preferences in the Mediterranean region(López-Martínez, 2017). In addition, studies have included assessments of the relationship between aesthetic qualities and landscape type(Wang et al., 2016) and agricultural landscape aesthetics(Zanten et al., 2016). These studies indicate that the visual quality in landscape design (Polat & Akay, 2015) is key to understanding how perception and cognition interact. However, research gaps still exist in how to apply these cognitive theories to actual landscape design practices and how to quantify the specific effects of different design elements on perception.

### 2.3.3 Unique Dynamics of Perception and Cognition in Cross-cultural Contexts (Taking Overseas Chinese Gardens as a Case Study)

This study aims to explore the interaction between perception and cognition in a cross-cultural context, taking overseas Chinese gardens as a case study. The cultural dependency of perception is reflected in the design elements of overseas Chinese gardens, which are deeply rooted in traditional Chinese culture and philosophy, and whose perception and appreciation are influenced by individual cultural backgrounds and cognitive frameworks(Goldstone et al., 2011). The contextual dynamics of perception points out that the perceptual experience of overseas Chinese gardens as a cultural symbol evolves as individuals deepen their understanding of Chinese culture(Brown et al., 2011). The interaction between cognitive style and environment points out that different cultural backgrounds foster different cognitive styles, which in turn affect individuals' perception and understanding of the environment. The cultural dissemination function of overseas Chinese gardens emphasizes their important role in cross-cultural communication and cultural diffusion, providing an ideal platform to facilitate the exchange of perceptions and cognitions among people from different cultural backgrounds(Mroczko-sowicz & Werning, 2012).

Therefore, this study focuses on exploring the perceptual and cognitive effects of overseas Chinese gardens in cross-cultural settings and how they serve as a medium for cultural exchange. This study examines how the public experiences and understands garden culture, and explores how garden design facilitates perceptual and cognitive exchanges in different cultural backgrounds. Overseas Chinese gardens, as sites of cultural communication in the context of globalization, provide a unique perspective for studying the effects of culture on cross-cultural perceptual and cognitive interactions.

### 2.4 Application of Large Language Models in Understanding Aesthetic Patterns

In the literature review of this study, we emphasize the background of rapid development and powerful application capabilities of large language models (LLMs), multi-agent systems (MAS), and AI agents in the field of artificial intelligence. MAS, as a core area of contemporary AI research, encompasses multiple interacting decision agents, with its vital role evident in distributed autonomous reasoning and behavior capabilities. LLM, with its efficient natural language processing capabilities, provides a new avenue for in-depth exploration of aesthetic cognitive patterns in overseas Chinese gardens. In the integration of virtual communities, these technologies not only enhance the quality of discussion and analysis but also make the exploration of aesthetic cognition more efficient and insightful. Through the fusion of these technologies, our research aims to explore aesthetic cognitive patterns, providing a dynamic platform for user interaction and insights.

## 3 methodology

### 3.1 analytical framework

This study takes to unfold according to the following experimental framework.

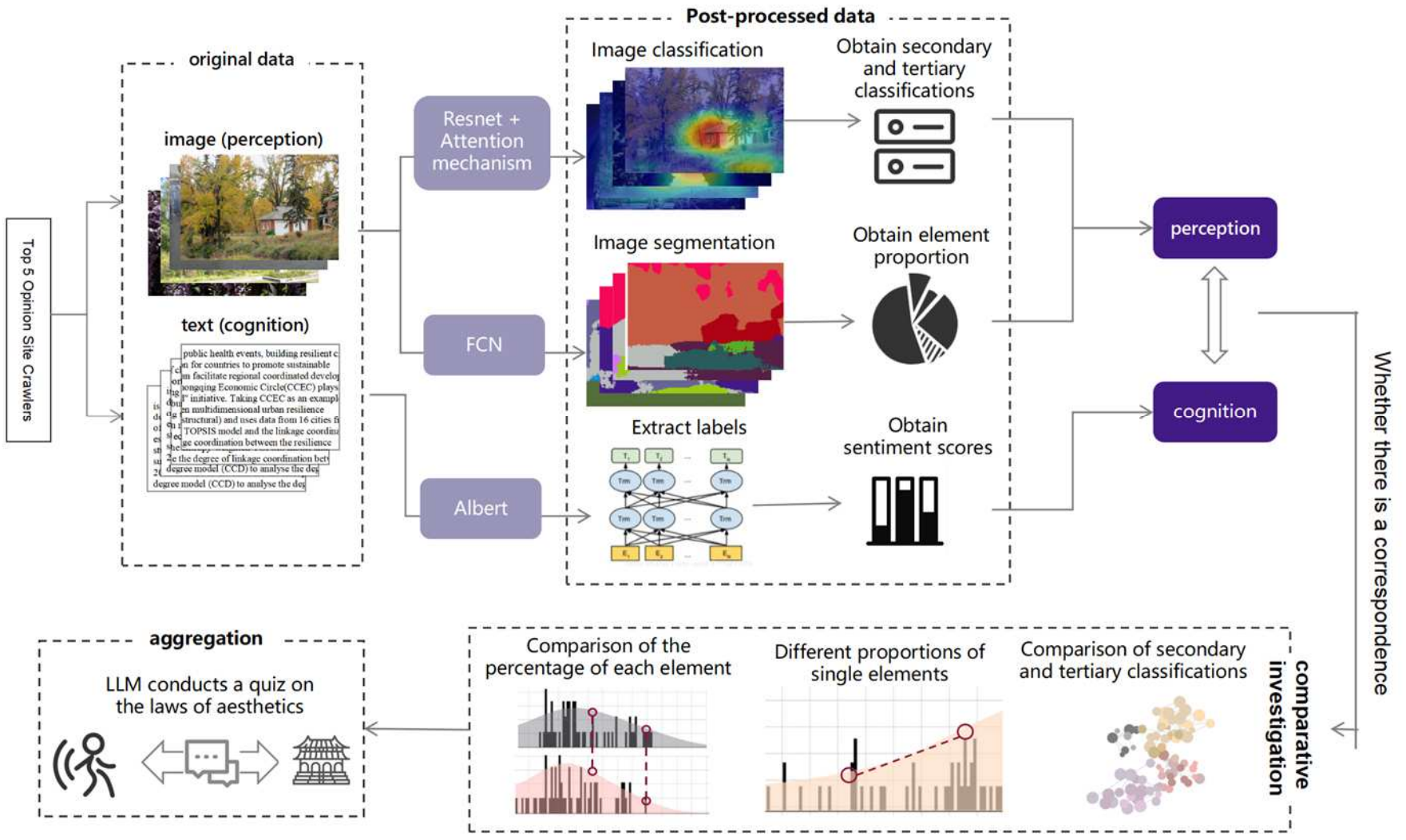

Figure 1 Analytical framework

### 3.2 data collection

In this study, we searched for comment threads with images from five foreign social media sites: Flickr, Twitter, Instagram, Tripadviser, and Reddit, using "Chinese Garden" as the keyword, and taking 2016-2023 as the time period. The time period from 2016 to 2023 was searched for comment threads with images. Among them, Flickr is a well-known photography sharing website; Twitter and Instagram are short text social media sharing websites with strong timeliness; Tripadviser is a well-known travel review website; and Reddit is a long text blog social media sharing website.

Each website has different information structure and different types. A total of 538,624 strip chart data are collected, constituting a multi-source heterogeneous network-wide perception database. Each piece of data includes a paragraph of text and a number of images, and each piece of data is linked by a unique ID to facilitate subsequent correlation analysis, as well as a number of other information types, as shown in Table 1.

| Categories | Information | Purpose |
|---|---|---|
| Raw Data Information | IDs | as digital credentials for all information linked to each piece of data |
| | Images | for post-excavation analysis of scene elements, Element proportion , and topic mode pattern |
| | Texts | for later mining of text keywords, text sentiment, etc. |
| | Additional comments | as supplementary data for text analysis |
| | Evaluation scores | to train models that predict the sentiment of text |
| | Number of likes (on a website) | as supplemental data for evaluation scores |
| | Publisher Identification Information | to distinguish between overseas and Chinese travelers |
| | Geographic Location | to distinguish between overseas Chinese gardens and local Chinese gardens |
| | Date | for filtering the time period of the data |

Tags for data classification training

Table 1 Types of raw data information

### 3.3 data preparation

The data processing stage is mainly divided into the following stages: 1. Image segmentation 2. Image preclassification 3. Image secondary classification 4. Sentiment score processing

### 3.3.1 Image segmentation

There is a corresponding image under each id, In this study, all images are subjected to image semantic segmentation with the help of a semantic segmentation model for urban images trained by China University of Geosciences based on the ADE_20K dataset and FCN full convolutional network(Lian Zhai, 2016).

The element segmentation map can be used as a scene mode pattern for each photo because the element segmentation map contains semantic information about different element locations and categories on the image. According to the image segmentation results, the scene elements and their proportion in each data can be obtained, as well as the scene pattern map. The final combing results that all the images contain 150 in scenes.

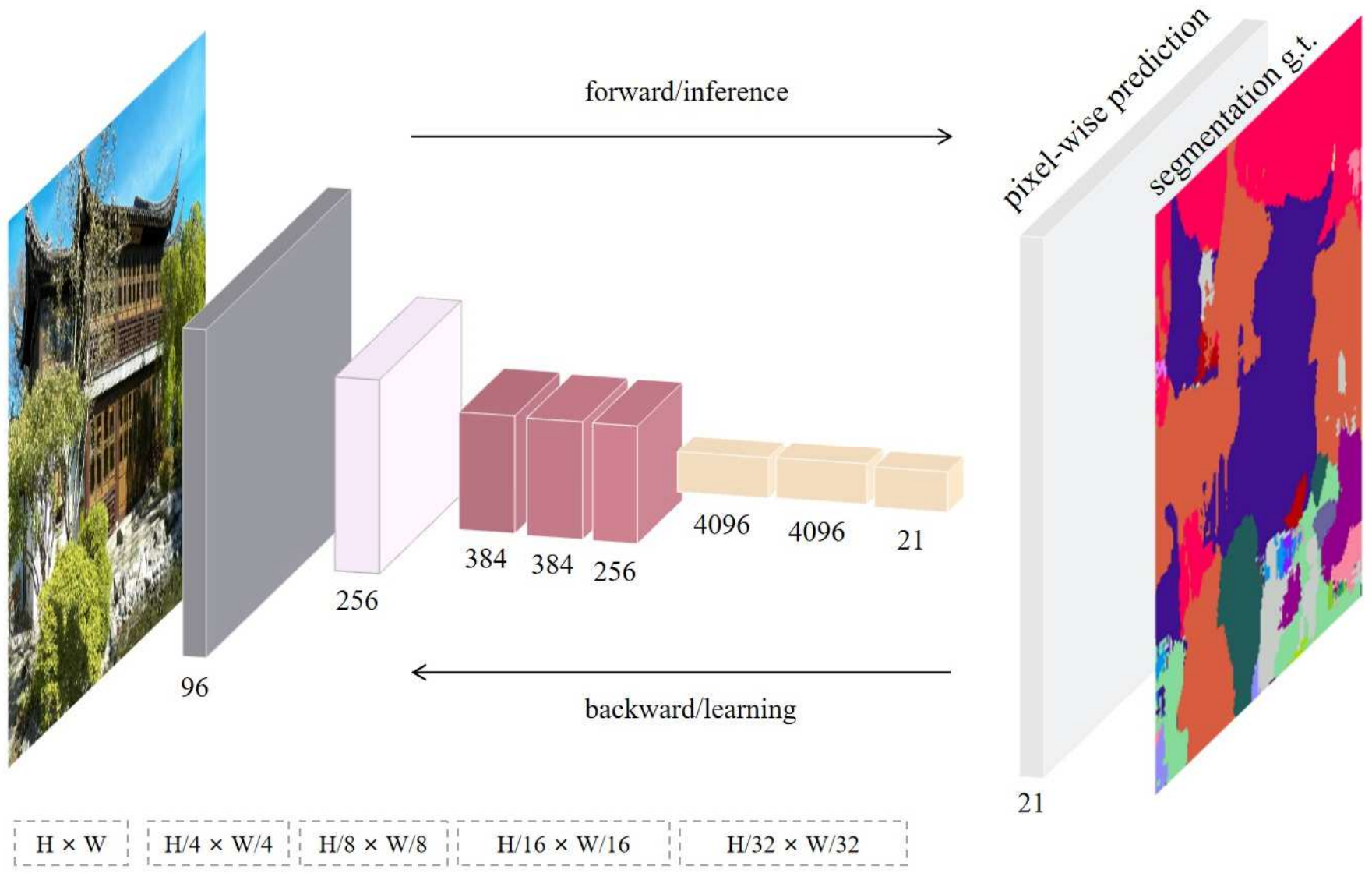

Figure 2 FCN schematic diagram

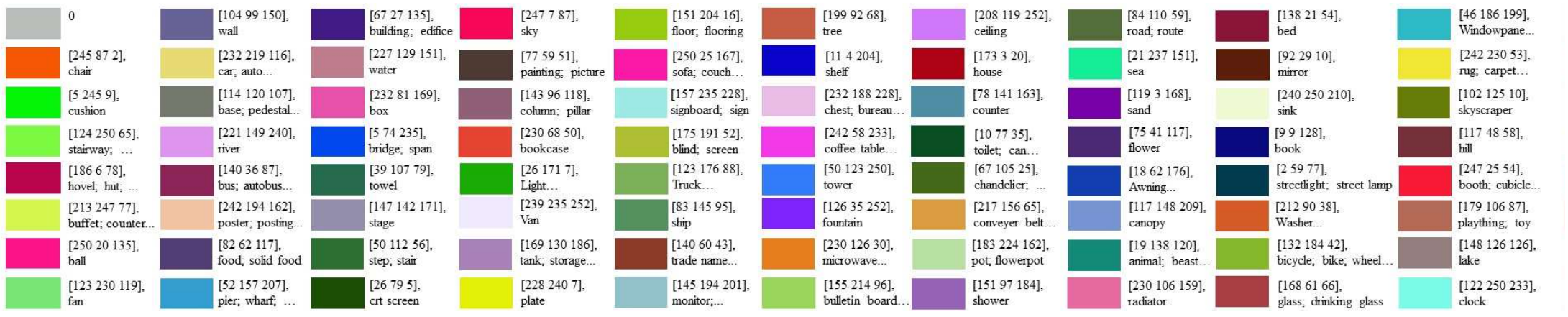

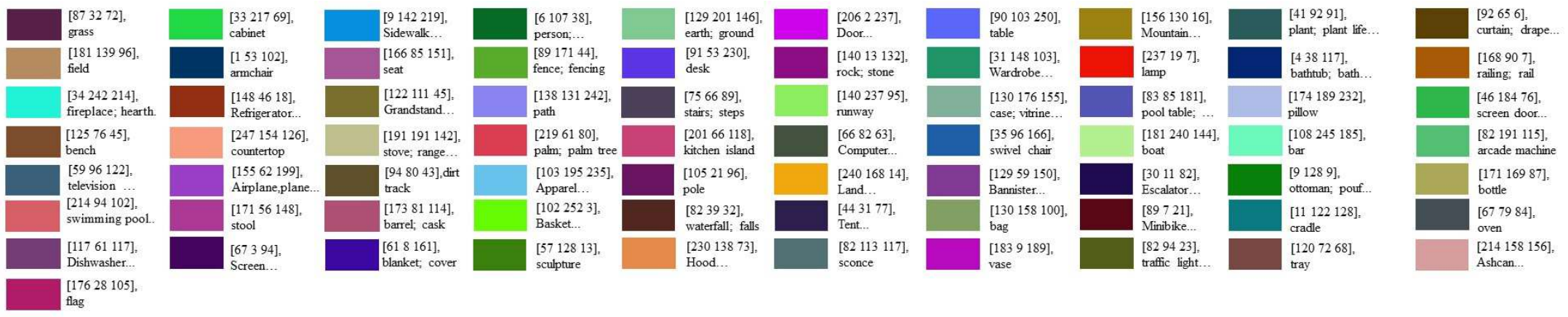

Figure 3 Segmentation legend

### 3.3.2 Image pre-classification

In this study, the image classification board, with the help of Place365 dataset developed by MIT(Zhou et al., 2017), combined with ResNet50 network and Attribute Mechanism (Attribute Mechanism) to simulate the human eye to observe the image (Nan, 2020; ZHU Zhangli, 2019), to get the keywords of the image, and also the weights of the different classification labels. Classification weights are divided into classification levels, the highest weight is the first level of classification, and so on three levels of classification, according to the observation that the weight of the image is less than 0.3 in the computer will cause difficulties in recognizing the difficulty of the image. Consequently, we directly discard third-tier image data falling below this threshold, retaining high-quality data. In addition to the text of non-English text data, the text word length is less than 8 is also deleted, so that the preclassification results of the data are obtained.

（1）Resnet

ResNet(He et al., 2015) is a deep learning architecture proposed in 2015 by Kaiming He et al. at Microsoft Research. The architecture solves the degradation problem in deep neural networks by introducing so-called "residual learning". In traditional deep networks, as the number of layers increases, the performance of the network tends to deteriorate due to gradient vanishing or gradient explosion. ResNet enables the network layer to learn the residuals between inputs and outputs by introducing residual blocks, i.e., $H(x) = F(x) + x$. This design allows the network layer to fit the residuals function directly, rather than expecting to fit a complete mapping, thus enabling the network to propagate gradients more efficiently through these direct connections. ResNet has been shown to perform well on a wide range of tasks and datasets, especially in the area of image recognition and classification, where its depth can scale to hundreds or even thousands of layers while still maintaining good performance.

（2）Attention Mechanisms

Attentional mechanisms(Vaswani et al., 2017) are a technique for modeling the focus of human attention in a neural network, which allows a model to focus on the most important parts of a large amount of input data as it is processed. The concept was first introduced in the field of natural language processing and has quickly been applied to a variety of different tasks and domains, including image recognition, speech recognition, and machine translation.

The core idea of the attention mechanism is that instead of being forced to process the entire input data at once, the network is able to learn the parts of the information that are most critical in a given task. This is usually achieved by assigning different weights or "attention", allowing the network to focus on the critical parts of the input data while ignoring less important information. The introduction of an attention mechanism has been shown to significantly improve the performance and efficiency of models, especially when dealing with complex or large-scale datasets.

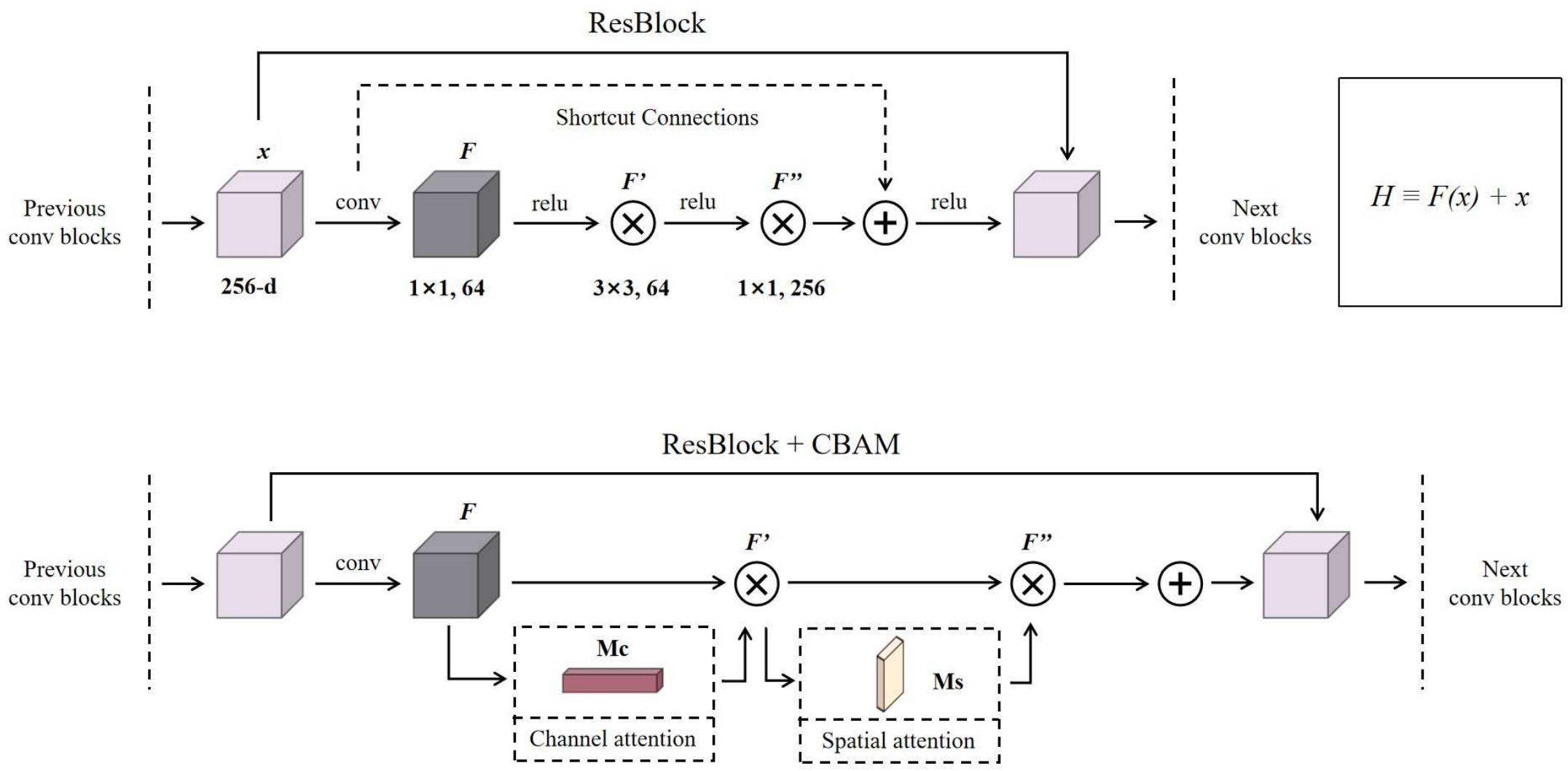

Figure 4 Diagram of resnet and resnet+ attention mechanism after schematic diagram (CBAM)

### 3.3.3 Secondary Classification of Images

In the pre-training phase, the ResNet deep learning model initially classified images of overseas Chinese gardens into a total of 365 categories. Integrated with specialized knowledge, we subsequently carried out a more detailed classification work. The work in this phase was not only based on the visual features of the images, but also considered in depth the cultural, historical background and artistic characteristics of overseas Chinese gardens. Combining the elements of garden design, style, materials used, vegetation types, architectural features, and other dimensions, these images were organized and classified into 17 major categories, 102 medium categories, and 139 subcategories.

Such a categorization not only reflects the rich diversity of overseas Chinese gardens, but also reveals their unique aesthetic and cultural values. This refined classification system helps to analyze and understand more deeply the visual presentation and cultural significance of overseas Chinese gardens on a global scale.

### 3.3.4 Sentiment Score Calculation (ALBERT) model

In this study, for the text information under each ID, we focus on extracting the sentiment scores. Using the 54,086 image-free text data in the Tripadviser dataset, combined with its star ratings, we train an ALBERT deep learning model to recognize the sentiment polarity of the text. This method effectively circumvents thesentiment recognition error due to linguistic features such as inflection and virtual voice compared to the traditional sentiment dictionary-based analysis(ZHAO HongYu & Abraham, 2018; Chenglong Zhang, 2018). To further reduce the potential error of deep learning, this study combines the sentiment dictionary to predict the sentiment polarity of each piece of data and calculate the sentiment score accordingly.

ALBERT (A Lite BERT) (Lan et al., 2019)is an advanced deep learning model designed for natural language processing (NLP) tasks. The model was developed by the Google Research team in 2019 to improve the performance and efficiency of the original BERT (Bidirectional Encoder Representations from Transformers) model. The main innovation of ALBERT is its parameter sharing mechanism and sentence order prediction improvement, which make it more efficient in dealing with large-scale datasets. Compared to BERT, ALBERT has a more streamlined structure that reduces the model size and training time while maintaining or even improving the task performance. The parameter sharing mechanism means that the same parameters are reused between different layers of the model, thus significantly reducing the total number of parameters of the model. In addition, ALBERT introduces sentence order prediction by modifying BERT's pre-training task, which contributes to a better understanding of contextual relationships in text.

## 3.4 MAS - AI Agents

### 3.4.1 Large Language Model

Large Language Models are deep learning algorithms designed for a variety of natural language processing tasks. These models are trained on massive datasets using transformer models, which gives them the ability to recognize, translate, predict, or generate text. The transformer architecture, comprising an encoder and a decoder, processes data by tokenizing inputs and applying mathematical equations to uncover relationships between tokens. This process, aided by self-attention mechanisms, allows LLMs to understand and generate human-like text. They consist of multiple neural network layers, including recurrent, feedforward, embedding, and attention layers, each contributing to the model's ability to capture the semantics of language and generate contextually relevant outputs.

### 3.4.2 AI Agents

AI agents are advanced software entities that perceive their environment, make decisions, and act on them to fulfill specific goals. Unlike traditional software, AI agents are proactive and capable of adapting to their environment. They leverage AI and machine learning algorithms to learn and evolve through interactions. These agents perceive digital surroundings using data inputs, make decisions by analyzing data, take actions based on these decisions, and continuously learn from the outcomes to improve their performance. This adaptive nature makes AI agents akin to intelligent assistants, capable of navigating and responding to the digital world dynamically and autonomously.

### 3.4.3 Multi-Agent Systems (MAS)

MAS are composed of autonomous agents interacting within a shared environment to achieve individual or collective goals. These systems are characterized by their agents' autonomy, interaction capabilities, goal orientation, and distributed control. Agents in MAS are capable of perceiving their environment, making decisions, and performing actions based on their internal algorithms or heuristic methods. The organization within MAS can vary, including flat, hierarchical, or networked structures, influencing how agents interact, cooperate, coordinate, negotiate, or compete with each other. MAS environments can be static or dynamic, accessible or inaccessible, deterministic or stochastic, and discrete or continuous, all of which affect the agents' interactions and behaviors.

### 3.4.4 Aesthetic cognition discussion community

This study utilizes a Multi-Agent System (MAS) to construct an artificial intelligence virtual community, aiming to explore the aesthetic cognitive patterns of overseas Chinese gardens. In this community, individual agents discuss in chat and combine networking with web searches in order to explore the scientific pattern of aesthetic cognition of overseas Chinese gardens.

The database constructed in this study, as well as the findings, serves as an external database to the community, providing a basis for discussion and serving as a monitoring tool.

This community, serves as a dynamic platform where researchers interact with user_agents to discuss specific research queries. As shown in Figure 5, researchers can design the data they want to analyze and define the various roles in the community.

Then, user_agents, supported by state-of-the-art large-scale language models, guide AssistantAgents in discussing and exploring the provided data to delve into the complexity of aesthetic perceptions of Chinese gardens. During the exploration process, these agents simulate real viewing scenarios and analyze the data, while they can network when necessary.

This approach not only utilizes the collective intelligence of different AIs, but also aligns with recent advances in multimodal deep learning, providing a nuanced understanding of aesthetic preferences. Furthermore, the combination of AI in this virtual discussion community ensures effective management and analysis of content, creating an environment conducive to authentic and insightful discourse. This innovative approach marks an important step in the exploration of aesthetic cognition, especially in the context of rich and diverse overseas Chinese gardens.

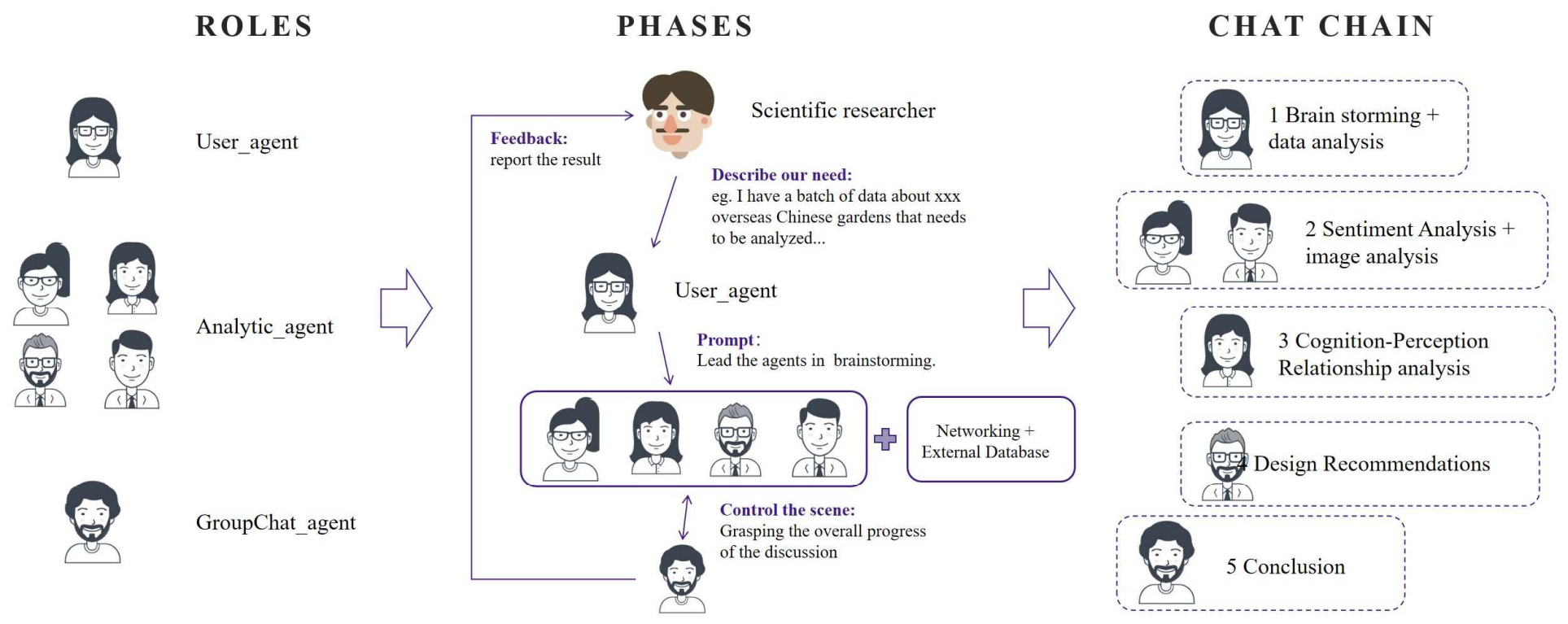

Figure 5 Division of labor process in the aesthetic cognition discussion community

## 4 Results and analysis

### 4.1 Image segmentation results

According to the image segmentation results, we can get the scene elements, the proportion of each data, and the scene mode pattern. The final combing results show that all the images contain 150 in the scenes.

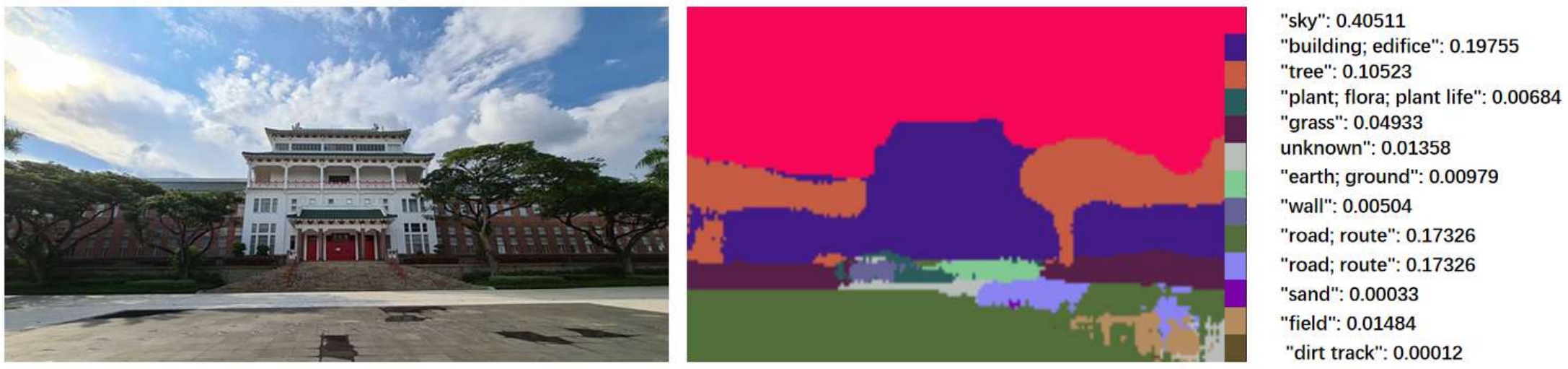

Figure 6 Segmentation result

## 4.2 Image classification results

In the pre-training stage, the ResNet deep learning model initially classified images of overseas Chinese gardens into 365 categories. Combining the elements of garden design, style, materials used, vegetation types, architectural features and other dimensions, we organize and classify these images into 17 large categories, 102 medium categories and 139 small categories.

In organizing the 17 major categories, we further divided them into aesthetic and functional spaces, presented in Figure 8 and 9 respectively. Functional space is mainly used for site infrastructure, so this study focuses on aesthetic space to investigate the relationship between perception and cognition.

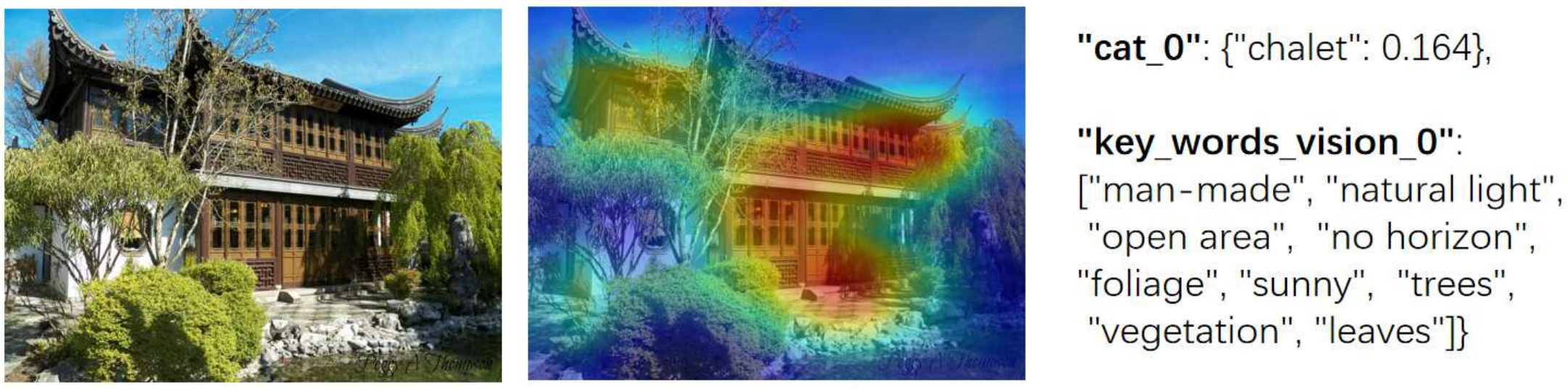

Figure 7 Resnet+ attention mechanism classification results

| | | |
|---|---|---|
| Commercial space | stall keeper | |
| | Formal dining room | 31;bakery, shop |
| | | 72;butchers_shop |
| | | 267;pizzeria |
| | | 284;restaurant; |
| | Specialty Chinese restaurant | 114;delicatessen; |
| | | 120;dining_hall; |
| | | 121;dining_room; |
| | Pub | 39;bar; |
| | | 54;beer_hall; |
| | utdoor b | 53;beer_garden; |
| | Tearoom | 322;sushi_bar; |
| | Coffee House | 99;coffee_shop |
| | | 185;ice_cream_parlor; |
| | Chinese Souvenir Shop | 160;general_store, indoor; |
| | | 162;gift_shop; |
| | | 146;flea_market, indoor; |
| Entertainment space | itern Fest | 7;amusement_park |
| | erforman | 16;arena, performance; |
| | ipatory a | 20;art_school; |
| | arts perf | 225;martial_arts_gym; |
| | tdoor pic | 265;picnic_area; |
| | Chinese | 286;restaurant_patio; |
| | reme Sp | 291;rope_bridge; |
| | tage Per | 315;stage, indoor |
| | Vater par | 353;water_park; |
| | oat on th | 277;raft; |
| | se templ | 335;toyshop; |
| Urban scene | City scene with | 47;bazaar, outdoor; |
| | | 319;street; |

Figure 8 Table Functional Scenario Classification Results

| | | |
|---|---|---|
| Garden borrowing | Moon gate | 12;arch; |
| | Long sea | 33;balcony, interior; |
| | Chinese | 37;bank_vault; |
| | Stained g | 90;church, indoor; |
| Single object | Aquatic a | 9;aquarium; |
| | Flowers | 141;field, wild; |
| | | 147;florist_shop, indoor; |
| | Pets | 201;kennel, outdoor; |
| | | 346;veterinarians_office; |
| | Sculpture | 236;museum, indoor; |
| | Insects | 239;natural_history_museum; |
| | | 342;underwater, ocean_deep; |
| | | 279;rainforest; |
| | Chinese | 297;science_museum; |
| | Wild anir | 341;tundra; |
| structure | Log cabin | 74;cabin, outdoor; |
| | | 299;shed; |
| | | 184;hunting_lodge, outdoor; |
| | Ordinary | 87;chalet; |
| | Ordinary Chinese pavilion | 159;gazebo, exterior; |
| | Sheds | 166;greenhouse, outdoor; |
| | ck and stone pagc | 214;lighthouse; |
| | | 334;tower; |
| | Chinese wooden pagoda | 251;pagoda; |
| | Traditional Chinese pavilion | 260;pavilion; |
| | Traditional large-scale Chinese architecture | 330;temple, asia; |
| | Modern abnormal-shape structures | 354;water_tower; |
| | Forest cottage | 339;tree_house; |
| Interior space | ttic and mezzanin | 26;attic; |
| | | 228;mezzanine; |
| | Ordinary door | 129;elevator, door; |
| | | 3;alcove; |
| | Chinese architecture interior main space | 331;throne_room; |
| | Wooden roof structure of Chinese architecture | |
| | Ordinary window | 123;doorway, outdoor; |
| | | 63;bow_window, indoor; |

| Traffic space | | | | | | | | | Open space | | | | | | | | Forest land | | |
|---|---|---|---|---|---|---|---|---|---|---|---|---|---|---|---|---|---|---|---|
| Wooden footway | Arch bridge | | | Roadway | Field avenue | Lane in forest | Boulevard | External stairway | Field | | | | Undeveloped wasteland | Lawn | | | bamboo forest | Broad-leaved forest | Mixed forests |
| 57;boardwalk; | 66;bridge; | 10;aqueduct,; | 347;viaduct; | 127;driveway; | 142;field_road; | 151;forest_path; | 152;forest_road; | 317;staircase; | 104;corn_field; | 138;farm; | 287;rice_paddy; | 1359;wheat_field; | 117;desert, vegetation,; | 164;golf_course; | 173;hayfield; | 209;lawn; | 36;bamboo_forest; | 150;forest, broadleaf,; | 338;tree_farm; |

| Borrow a scene outside the garden | | | | | Stone skits | | | | | |
|---|---|---|---|---|---|---|---|---|---|---|
| Mountain | | | Cliff | City scene overview | stone carving | | Stone tablet | Cave | Rockery | Rock layout |
| 73;butte; | 81;canyon; | 344;valley; | 94;cliff; | 348;village; | 69;burial_chamber; | 13;archaelogical_excavation; | 86;cemetery; | 167;grotto; | 289;rock_arch; | 292;ruin; |

| Water body | | | | | | | | | | | | | | | | | | | | |
|---|---|---|---|---|---|---|---|---|---|---|---|---|---|---|---|---|---|---|---|---|
| Waterfront architecture | | | Waterfront space | Forest streams | Cascade | | Lotus pond | Sculpture Fountain | Natural stone pool | Large-scale lake | Large-scale lake dotted with buildings | Tiny pier | | Wetland | | Natural pool | | Waterfront corridor | Ditch | Waterfall |
| 59;boathouse; | 229;moat, water,; | 79;canal, urban; | 78;canal, natural,; | 110;creek; | 113;dam; | 357;wave; | 145;fishpond; | 154;fountain; | 180;hot_spring; | 204;lagoon; | 205;lake, natural; | 218;lock_chamber; | 266;pier; | 224;marsh; | 323;swamp; | 271;pond; | 324;swimming_hole; | 273;promenade; | 340;trench; | 355;waterfall; |

| Narrow space | | | | Snow scenery | | | | | Courtyard space | | | | | | | | | | | | | | |
|---|---|---|---|---|---|---|---|---|---|---|---|---|---|---|---|---|---|---|---|---|---|---|---|
| Alley | Chinese pergola | | Indoor corridors | Animals in snowfield | Recreation in snowfield | Buildings in snowfield | | Snowfield | Pavage in courtyard | Outdoor lounge | Plant garden | Chinese architectural courtyard | Flower garden | Stone courtyard | Flowering shrubs | Greenhouse | Orchard | Green leafy woody plants | Vines | Green sculpture | | Large-scale flower beds |
| 4;alley; | 11;arcade; | 272;porch; | 106;corridor; | 186;ice_floe; | 189;ice_skating_rink, outdoor,; | 191;igloo; | 304;ski_resort; | 309;snowfield; | 109;courtyard; | 259;patio; | 362;yard; | 197;japanese_garden; | 290;roof_garden; | 364;zen_garden; | 62;botanical_garden; | 165;greenhouse, indoor,; | 249;orchard; | 345;vegetable_garden; | 349;vineyard; | 333;topiary_garden; | 153;formal_garden; | 140;field, cultivated,; |

Figure 9 Table Classification results of aesthetic scenes

## 4.3 Secondary classification analysis

### 4.3.1 Overall analysis

In this study, we conducted an in-depth analysis of types of space and natural landscape elements in overseas Chinese gardens to explore the public's perceptions and emotional responses to these elements.

We found that in terms of the perception of types of space, the public's perception of different types of space (including interior space, traffic space, narrow space, courtyard, and empty space) showed a clear tendency to concentrate, suggesting that types of space play a key role in shaping people's perceptual experience.

In terms of the perception of natural elements, we analyzed the perception of woodland, botanical gardens, water bodies and snow scenes. We found that the frequency of these elements was high. Particularly striking was the perception of water bodies, which was significantly higher than the other types in terms of the number of photographs captured and the number of comments (Figure 10), revealing the public's high sensitivity to water features.

For the perception of water bodies, we plan to conduct a more in-depth analysis in the subsequent chapters to explore their status in the public's mind and the emotional responses they evoke. We will analyze visitors' sentiment scores when they see different water feature modes. This analysis will help us understand water bodies' unique status and cultural significance in overseas Chinese gardens.

In addition, we also examined the perception of architectural elements specific to Chinese gardens. In terms of affective ratings, people showed a significant trend of single-peak responses to these architectural features, and they generally exceeded 50%. Among them, the plurality of sentiment scores for Chinese wooden pagodas, traditional Chinese pavilions, and traditional large-scale Chinese buildings were all centered at 0.7-0.8 (Figure 10), demonstrating the significant aesthetic and emotional appeal of these architectural features, thus complementing our understanding of the aesthetic and cultural value of overseas Chinese gardens.

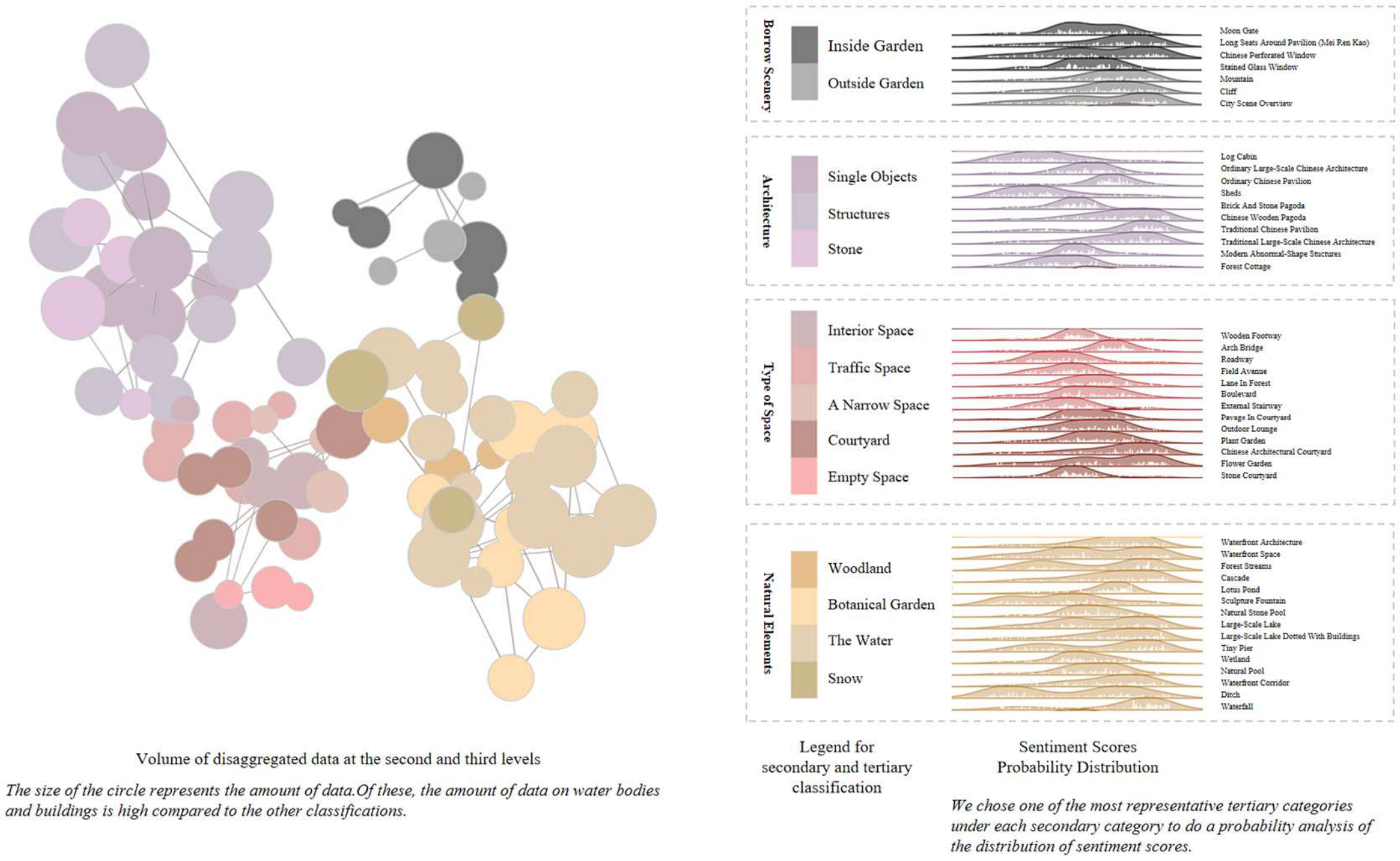

Figure 10 Percentage of each classification

**4.3.2 Typical Case - Water Feature Focus Analysis**

In categorizing water features in overseas Chinese gardens, we subdivided them into four main categories: regular water features, natural water features-lakes, natural water features-streams, and waterfront buildings. According to Figure 11, for regular water features, the statistical analysis of emotional probability distribution revealed that the emotional responses were relatively uniform and low, which indicated that regular water features did not elicit strong emotional responses. Regular water features appeared more subdued regarding emotional arousal than the other categories.

Further examining the naturalistic water feature-lake category, including naturalistic pools, naturalistic lakes, large lakes, and large lakes with architectural accents, we found that these water features elicited stronger and generally positive emotional responses. This finding contrasts with regular water features and reveals the effectiveness of naturalistic water features in eliciting positive emotions.

The distribution of emotions showed a bifurcation for the naturalistic water feature-stream category. Among them, due to their landscape richness, streams tended to have a more positive emotional response overall, showing a significant difference from the lake category. This result highlights the influence of the diversity of landscape elements on emotional responses.

When analyzing the category of waterfront architecture, we found that its emotional responses were the most positive of all the categories, and the data showed that waterfront architecture showed a single peak. The highest points were all greater than 0.6 points. This result contrasts with naturalistic water features and shows the significant role of architectural elements in emotional arousal.

Considering the emotional distributions of these different water feature categories together, it is clear that there is a significant difference in emotional responses between naturalistic and regular water features, with naturalistic water features tending to elicit stronger and more positive emotions.

In addition, waterfront architecture showed the highest positivity regarding emotional responses due to its unique visual and spatial characteristics. This study provides insights into understanding the effects of different water feature categories on emotional responses in overseas Chinese gardens by analyzing these data. It provides crucial theoretical support for water feature design's emotional impact and aesthetic role.

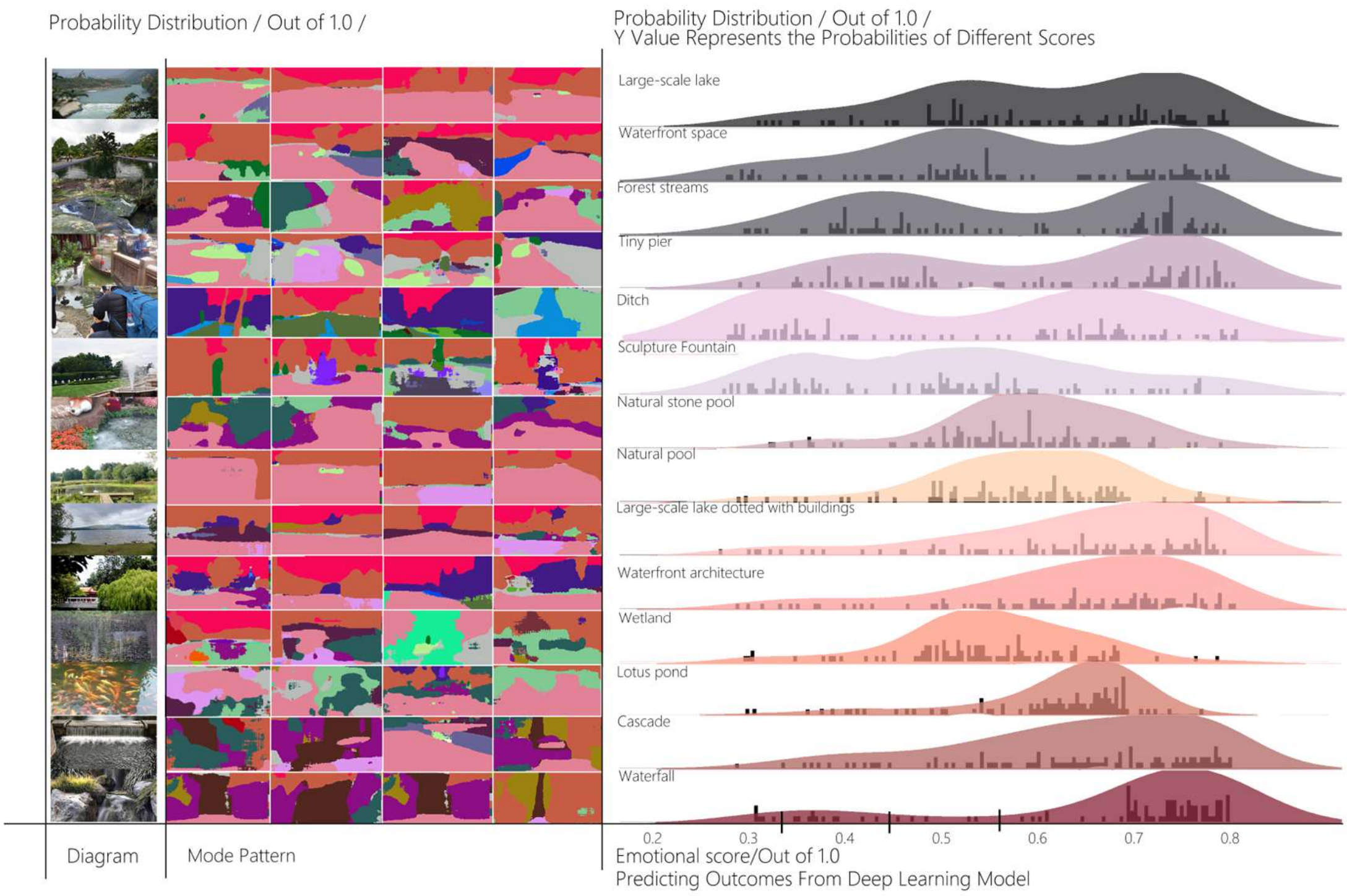

Figure 11 Analysis of the water feature

## 4.4 Three-level classification analysis

In the detailed analysis and comparison of the various subcategories of overseas Chinese gardens, we find that compared with plants, scenes with Chinese architectural features are the distinctive characteristics of these gardens. Therefore, the next section of this paper will focus on discussing and analyzing "scenes with Chinese architecture". In order to deeply understand the perceptual and cognitive impacts of these scenes, we will take a specific approach to establish the connection between scene segmentation maps (representing perception) and emotion text distribution (representing cognition).

The essential elements of traditional Chinese gardens include pavilions, terraces, buildings, pavilions, Xuan, salons, corridors, and boats, often found in overseas Chinese gardens. Based on the characteristics of garden scenery, we analyze them into two categories. First, based on the scenes containing pavilions, pavilions and Xuan, we will analyze the influence of the proportion of these perceptual elements on cognitive emotions. Second, for scenes containing corridors and bridges, we will explore the role of the openness perception of space on cognitive emotions.

### 4.4.1 Effect of the ratio of perceptual elements on cognitive emotions

Landscape architecture containing Chinese elements is an essential feature of overseas Chinese gardens. Since the different angles of the photographer will affect how the actual building size is reflected in the photographs (Xiang, 2008)[42], the proportion of plants divided by the proportion of buildings was used in this study (Figure 12) to analyze the relationship between the proportion and the sentiment.

From the distribution of emotions, it can be seen that when the proportion of plants to buildings is close to 2:1, the distribution of emotions is concentrated in the positive range (greater than 0.6), which to some extent reflects the theory of "three parts of water, two parts of bamboo, and one part of the house"(Qi, 2019) [43] described by Li Gefei. It also confirms that although foreign tourists have a low cognition of Chinese garden culture, their aesthetics can conform to the general design laws.

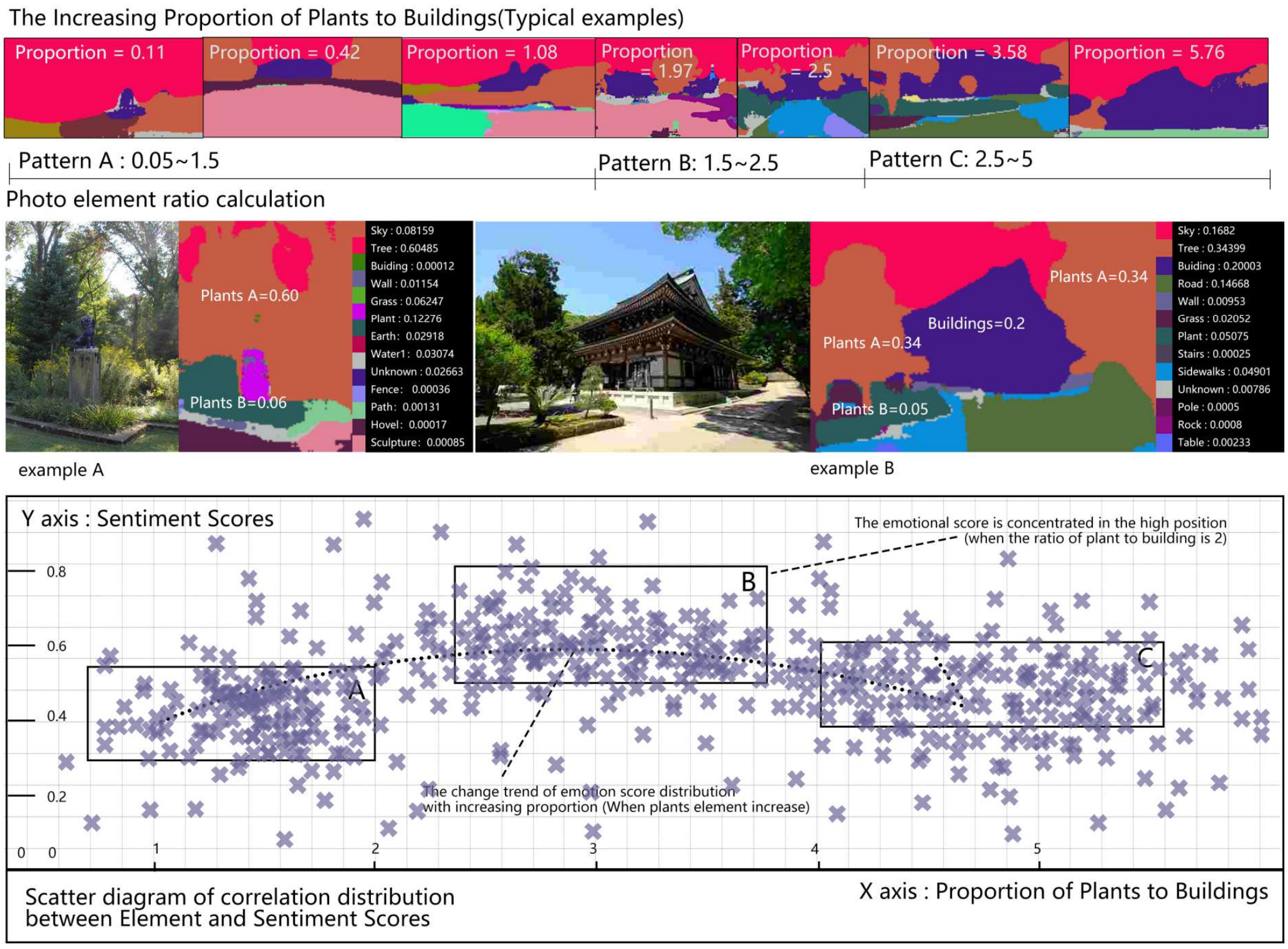

Figure 12 Analysis of proportion of plants to buildings

Furthermore, simply two elements are still needed to explain the problem entirely. Comparing the characteristics of multiple elements can more accurately respond to the aesthetic laws of foreign tourists. As shown in Figure 13, the pictures under the subcategory of the pavilion are traversed to call out the four elements of water, plant, building, and stone, according to which they are categorized into three types: building+plant, building+plant+stone, building+plant+stone+water. Plotting the distribution of the emotions of these three types of scenes stacked on each other, We found that scenes with all four elements have the highest scores, whereas the distribution of scenes with and without stones is approximate but higher overall. Having all the elements is another vital factor in attracting the attention of tourists, with water bodies having a more significant impact.

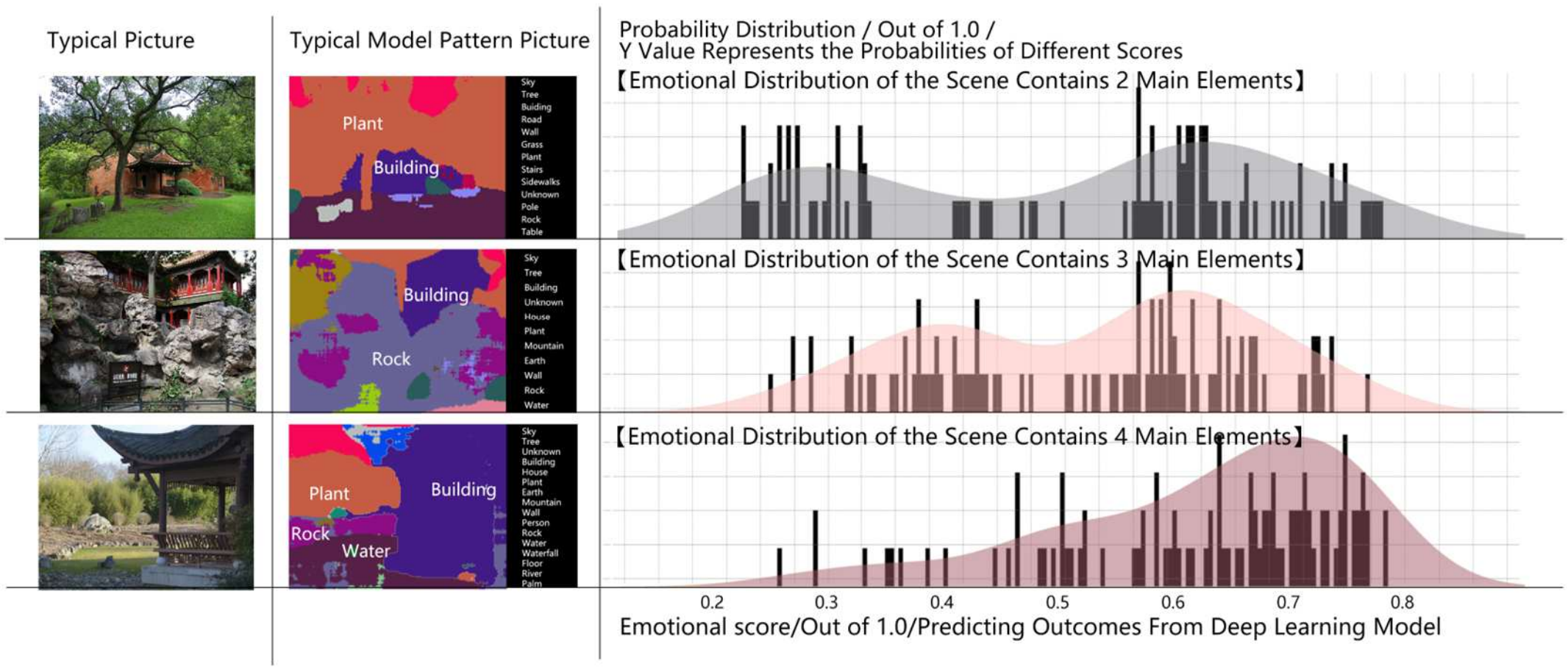

Figure 13 Analysis of multiple elements

### 4.4.2 The effect of spatial openness on emotions

For the effect of spatial openness on mood, we chose a landscape scene containing a gallery. In the landscape of corridors and walls, unique windows are also important to express the mood of Chinese gardens.

In this study section, we chose the percentage of elements related to spatial division, such as walls. Considering that wall elements usually include walls on both sides of the space, we divide their percentage by two to reflect their proportion more accurately.

As shown in Figure 14, through the scatterplot of wall occupancy versus mood distribution, we found that the mood scores decreased significantly when the wall occupancy exceeded 0.2, and the data showed a concentration of 0.1-0.3 points. This finding suggests that overly enclosed spaces may hurt people's emotions.

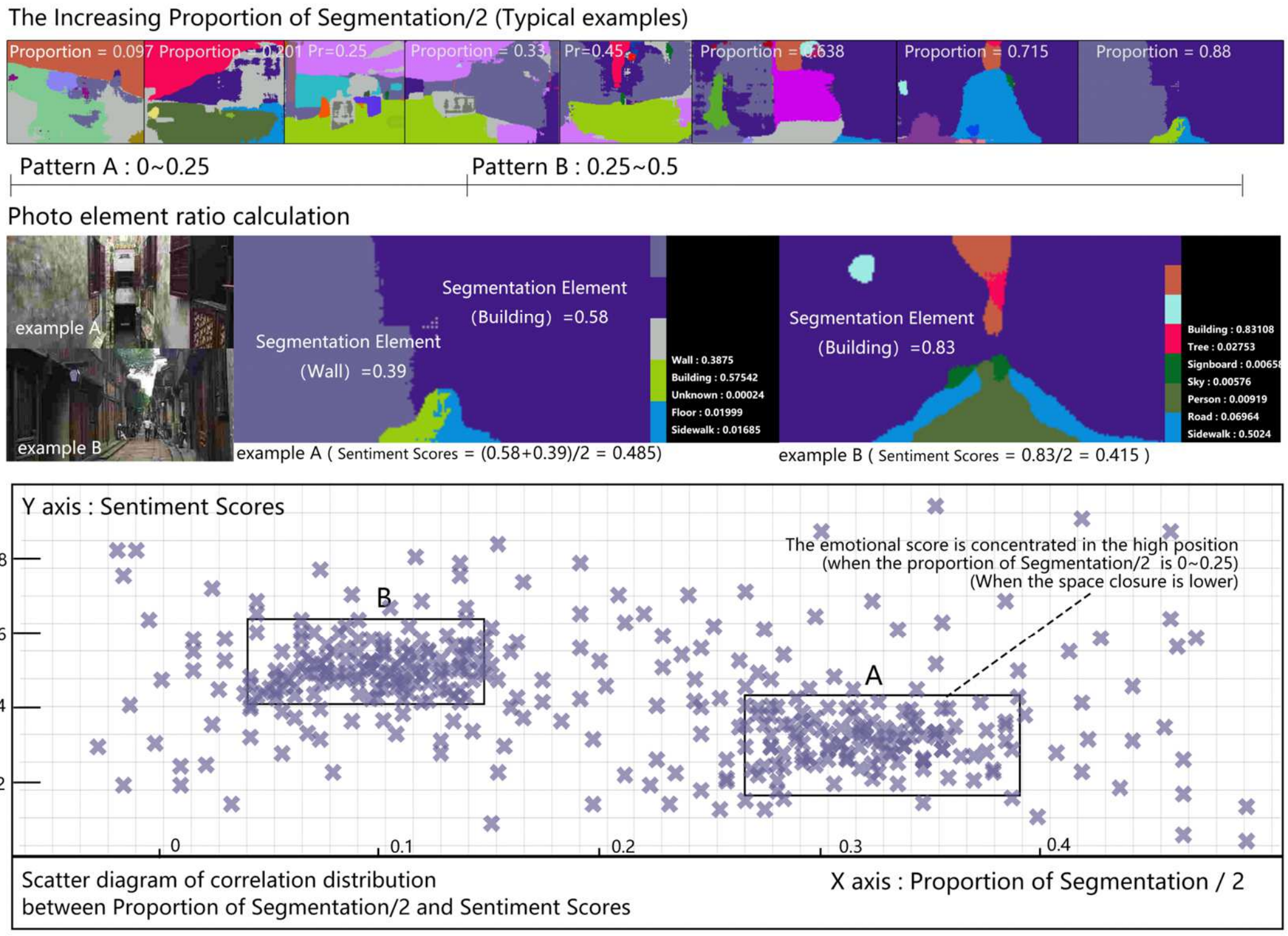

Figure 14 Analysis of proportion of segmentation

Next, we conducted an in-depth comparative analysis of the narrow space subcategory with the porch shelf subcategory and compared the leaky window subcategory with the window subcategory with each other. The study results in Figure 15 showed that the mood distributions of the porch shelf and leaky window categories generally showed a more positive trend. Specific data indicated that both the porch frame and leaky window categories showed a single peak in the sentiment analysis, with the porch frame category's sentiment scores clustered between 0.5 and 0.6, while the leaky window category's sentiment scores were clustered between 0.6 and 0.7. This finding emphasizes the important role of design elements such as leaky windows and porch shelves in expanding the perception of space and guiding visitors to experience positive emotions. In contrast, the peak emotions of the closed alley category were concentrated around 0.25 points, with an overall tendency towards negativity, suggesting that scenes lacking Chinese garden features may not be effective in triggering positive emotional responses from visitors.

In addition, the analysis of the closed window category revealed a striking double-peak phenomenon. In the negative region, its emotional score was around 0.3; in the positive region, it appeared around 0.7. This double peak in the sentiment distribution reflects that window designs with Chinese characteristics trigger different reactions among foreign tourists. If tourists could recognize these elements as characteristic of Chinese culture, they tended to display positive emotional responses. However, those tourists who fail to recognize this cultural feature may misinterpret them as merely closed scenes, resulting in a negative emotional experience. This result provides practical guidance for designing garden spaces and insight into the complexity of how overseas Chinese gardens interact with visitors in a cross-cultural context. This insight is essential for understanding and designing garden spaces that enable cross-cultural communication, emphasizing the need to consider cultural recognition and cultural education in the design process.

In summary, the pattern of overseas tourists' aesthetics of Chinese gardens conforms to the general pattern, which reinforces the above characteristic of overseas tourists' low level of awareness but high acceptance of Chinese gardens. Their evaluation of the design work comes from the quality of the scene design rather than the Chinese elements, and a scene with rich and harmonious elements can better arouse aesthetic pleasure.

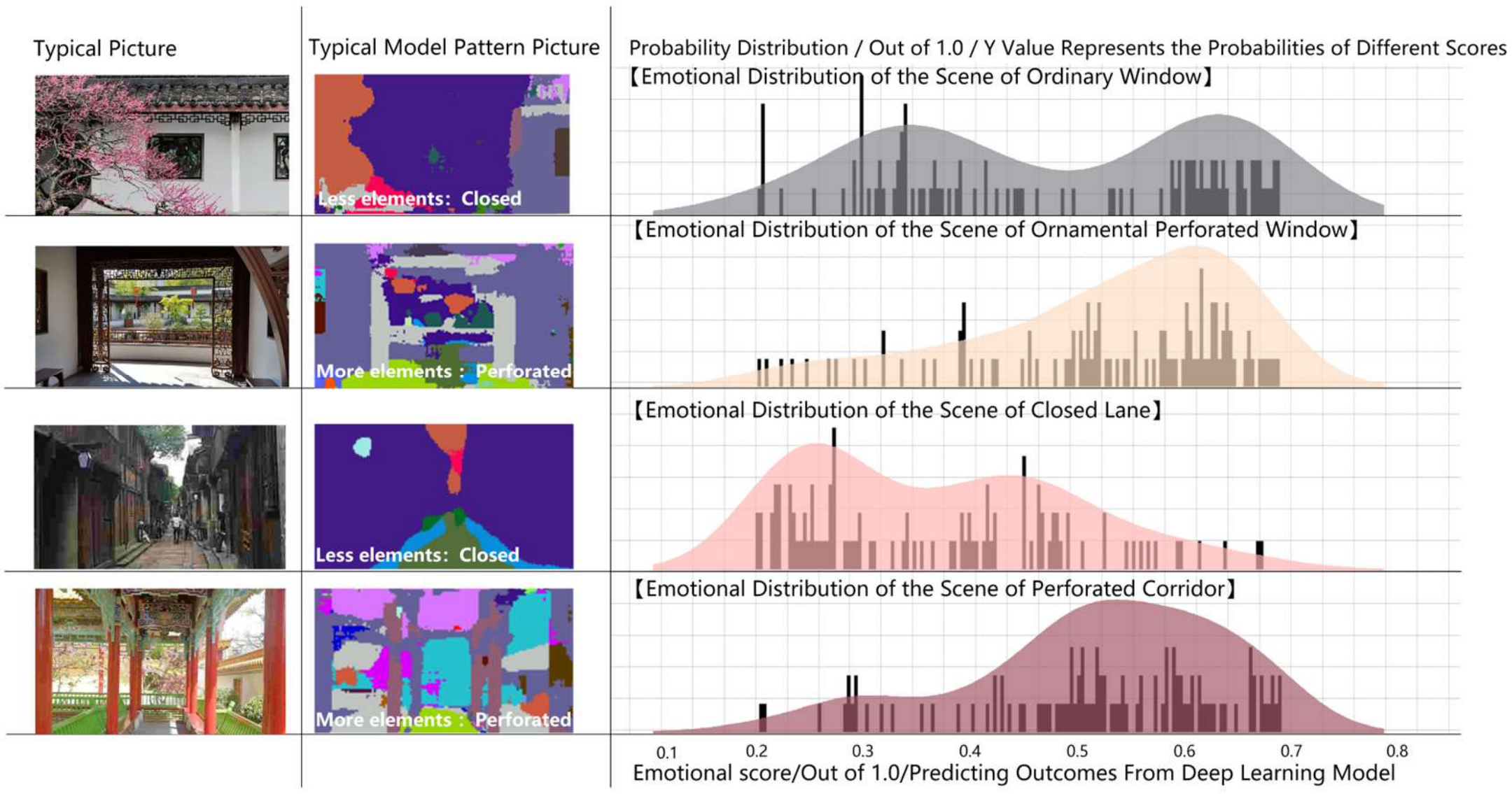

Figure 15 Analysis of elements of the narrow space

## 4.5 AI Agent

In this study, researchers input the "Urban scene. json" file, which contains data related to various urban scenes, including sentiment scores, image segmentation, and visual keywords. Agents can comprehend this information and respond accordingly (Figure 16).

AI agents analyze the emotional scores and image information separately, using this analysis to infer the relationship between tourist perception and cognition. Based on these insights, targeted design recommendations are formulated. Through discussion and refinement, these agents are capable of correlating multimodal and diverse information for joint analysis. They progressively deduce and explore internal connections and developmental patterns. For instance, by linking the sentiment score of "moderately positive" with the presence of "Chinese elements in urban environments" in corresponding images, they arrive at the conclusion that people perceive aesthetic appeal when familiar urban settings incorporate Chinese elements. By combining the sentiment score analysis of "natural elements" with the image analysis, the design suggestion of "balancing man-made elements with more natural elements, such as trees or bodies of water," is given. The insights from the textual sentiment analysis were combined with the actual spatial experience design practice.

Agent's response involves an analysis of the proportion of a single classification, focusing on the effectiveness of natural water features and traditional Chinese architecture in evoking positive emotions. Furthermore, the agents divide the proportion of plants by the proportion of buildings, deeply exploring the relationship between various element proportions and sentiment. Ultimately, they conclude that balancing artificial structures with natural elements and cultural features can further enhance the perception and sentiment responses, completing the analysis of the impact of the proportion of perceptual elements of cognitive sentiment. This demonstrates the effectiveness of our external knowledge repository.

**Aesthetic Cognition Discussion Community**

**Scientific researcher:**
I have a batch of data about xxx overseas Chinese gardens that needs to be analyzed ...

**User_agent:**
Based on the analysis of the "Urban scene.json" file, we can gain insights into the aesthetic cognition and perception of public viewing overseas Chinese gardens in urban scenes. This file contains data on various urban scenes, including detailed information about sentiment scores, image segmentation (seg), and vision keywords.

**Analytic_agent:**
**Sentiment Analysis**:
The sentiment scores in the data represent the emotional response of the public to these scenes. The sentiment scores are consistently around 52-55, indicating a moderately positive emotional response. This suggests that the public generally has a favorable view of these urban scenes, possibly due to the presence of Chinese elements in a familiar urban setting.

**Image Segmentation (seg) and Vision Keywords:**
The 'seg' data shows a significant presence of elements like buildings, sidewalks, roads, and people in these scenes. This indicates that the scenes are bustling urban areas with active street life.
The vision keywords like "man-made," "natural light," "open area," "driving," and "asphalt" suggest an urban environment that is both man-made and open, with significant vehicular movement. These elements contribute to the perception of these areas as functional and vibrant urban spaces.

**Cognition-Perception Relationship:**
The relationship between cognition (emotional response) and perception (image data) in this context can be understood as the interplay between the emotional appeal of these scenes and their visual characteristics. The moderately positive sentiment scores are likely influenced by the perception of these scenes as lively, functional, and possibly aesthetically pleasing due to the integration of Chinese elements in an urban setting.

**Design Recommendations:**
To enhance the public's emotional scores in such scenes, it might be beneficial to balance the man-made elements with more natural features like trees or water bodies, as these can create a more soothing and visually appealing environment.
Incorporating more Chinese architectural styles or decorative elements in buildings could strengthen the cultural appeal and uniqueness of these scenes.
Improving pedestrian areas and reducing vehicular dominance could enhance the public's comfort and enjoyment of these spaces.

**GroupChat_agent:**
Overall, the data from "Urban scene.json" suggests that the public responds positively to urban scenes with Chinese elements, appreciating both their functionality and aesthetic appeal. Balancing man-made structures with natural elements and cultural features could further enhance this positive perception and emotional response.

Figure 16 Analysis results from the Aesthetic Cognition Discussion Community

## 5 discussion

This study has significant theoretical and applied implications in the field of perception and cognition.

First, from the theoretical level, by comprehensively analyzing social media opinion data, this paper successfully reveals the relationship between scene categories and emotion distribution in overseas Chinese gardens, as well as the relationship between specific scene elements and emotion distribution. This finding not only enriches the theoretical framework of perception and cognition interaction research, but also provides a new perspective for understanding the impact of cultural landscapes on emotion and cognition.

Second, the research methodology in this paper shows innovation in the application of technology. Our proposed analysis method based on social media data, combined with multimodal techniques and the latest LLM as well as AIAGENTS technologies, opens up new paths for future research in the fields of cultural landscape, spatial design, and emotional cognition.

### 5.1 Cultural and Aesthetic Values of Overseas Chinese Gardens

The results of this study reveal the unique cultural and aesthetic value of overseas Chinese gardens. Despite the fact that the number of overseas Chinese gardens is considerable and in the developmental stage, overseas tourists are relatively immature in their knowledge of Chinese gardens, often out of curiosity and often confused with Japanese gardens. This shows that although the form of Chinese gardens has been spread and accepted to a certain extent overseas, their deeper cultural and aesthetic values, such as the spiritual connotations of "heaven and earth in a pot", "mountains and forests at one's fingertips", and "the idea of seclusion", have not been recognized by overseas tourists. However, its deeper cultural and aesthetic values, such as "the world in a pot", "mountains and forests at your fingertips", and "the idea of seclusion", have not been widely understood.

In addition, the detailed analysis of the aesthetic scenes shows that the preferences of overseas tourists for Chinese gardens are not significantly different from the aesthetic rules of gardens in general. They place a high value on the architectural elements in Chinese gardens, while their fondness for natural elements such as water features and plants stems mainly from the pursuit of natural beauty rather than an in-depth understanding of the unique aesthetics of Chinese gardens. This finding suggests that although overseas tourists have a certain degree of acceptance of Chinese gardens, their knowledge of their unique cultural and aesthetic values is still at a shallow stage.

In summary, although overseas Chinese gardens have achieved some success in cultural dissemination, the dissemination and understanding of their deeper cultural and aesthetic values still need to be strengthened. Future work should focus on enhancing the depth of overseas tourists' knowledge of the unique cultural and aesthetic values of Chinese gardens in order to promote cultural exchange and understanding in the true sense.

### 5.2 Implications of Multimodal Techniques and AI Agents for Aesthetic Cognition Research

The multimodal data analysis approach adopted in this study significantly enhances the in-depth exploration of aesthetic cognition. Traditional web-based cognitive research usually relies on textual data, which limits the comprehensive understanding of public cognition, especially in revealing the perceptual patterns behind cognition. Our approach combines public opinion images with textual data, enabling direct correlation of scene perception and emotional response, and a deeper understanding of the impact of design forms on user perception. For example, the relationship between the percentage of architectural and botanical elements and emotional scores is analyzed, revealing the impact of different combinations of design elements on tourists' preferences. This method not only enhances the cognitive research methodology based on network data, but also helps to dig deeper into the perceptual laws behind public cognition, providing theoretical support for urban planning and design.

In addition, the method of combining big data and multimodal technology brings breadth to aesthetic cognition research. In the research field of aesthetic cognition, traditional methods rely on converting individual perceptual experiences into emotional scores, and then analyzing the relationship between these quantitative data. While this approach is effective to a certain extent, it has limitations in capturing complex emotions and cognitive details. In contrast, the inclusion of AI agents has revolutionized aesthetic cognition research. The introduction of AI agents is not just a technological innovation, but represents a methodological leap. By directly analyzing textual data, researchers can more deeply understand and interpret the emotional and cognitive structures embedded in opinion data.

The ability of AI agents to process and parse textual data allows research to go beyond simple digital analysis. For example, through natural language processing techniques, AI agents can identify emotional coloring, semantic associations, and even reveal implicit cultural and social contexts in texts. This multi-dimensional analysis method not only improves the depth of the research, but also broadens its breadth. For example, when analyzing the aesthetic perceptions of overseas Chinese gardens, AI agents can help identify and explain the aesthetic differences of viewers from different cultural backgrounds, revealing the commonalities and differences in the aesthetic perceptions of different social groups.

In addition, AI agents also excel in the efficiency and accuracy of data analysis. While traditional manual analysis methods are not only time-consuming but also susceptible to subjective factors, AI agents can process large amounts of data quickly and accurately, ensuring the objectivity and reliability of the research results. This is essential for understanding and interpreting complex patterns in large-scale data sets.

In summary, the incorporation of AI agents has not only achieved significant enhancement in the depth and breadth of aesthetic cognition research, but also significantly improved the efficiency and accuracy of data analysis. This methodological innovation opens up new paths in the field of aesthetic cognition research and heralds the further development of theory and practice in the field. Looking ahead, it is reasonable to assume that AI agents will play a central role in analyzing complex aesthetic phenomena, especially at the level of cross-cultural and multimodal data analysis, and the application of AI agents in this field will provide a more in-depth and comprehensive understanding of the study of aesthetic cognition, and promote breakthroughs and innovations in both academia and practice.

### 5.3 Practical application implications

The results of this study have important implications for architectural and landscape design practice, especially in promoting cultural diversity and cross-cultural communication. In future construction, the design of Chinese gardens as cultural symbols is not limited to the creation of space, but should also focus on the display and promotion of cultural connotations. Based on an in-depth understanding of tourists' preferences, designers can consider implementing "soft" cultural construction by combining tourists' attraction to specific elements. For example, around the typical garden buildings, cultural activities closely related to Chinese garden culture can be organized, so that these buildings become the core symbols of cultural dissemination. In addition, the content deployed alongside water bodies can reinforce the cultural connotation of garden hydraulic design, for example, through interpretive signs or interactive exhibitions about the philosophy and technology of water in Chinese gardens.

The exploration of aesthetic cognitive laws in this study provides guidance for more effectively integrating the spirit of Chinese garden culture into the gardening experience. Designers can utilize these cognitive laws to plan spatial layouts, select materials, and even design gardening routes to ensure that visitors not only visually enjoy beautiful landscapes, but also have a deeper experience on a cultural and emotional level. For example, visitors can be guided through different landscape elements through well-designed paths, thus invisibly conveying the philosophical and artistic concepts of Chinese gardens.

In addition, these findings can help promote understanding and appreciation of Chinese garden culture on a global scale. By considering the preferences and ways of perception of visitors from different cultural backgrounds, designers can create more inclusive and engaging spaces that promote cultural exchange and understanding. This approach not only enhances the visitor experience, but also provides new avenues for cultural preservation and dissemination. Therefore, the results of this study are not only important for theoretical research, but also provide practical guidance for actual architectural and landscape design.

### 5.4 Research limitations and future research directions

This study has achieved some results in exploring the perceived and cognitive interactions of overseas Chinese gardens, however, several limitations still exist. First, the sample data sources of this study are mainly limited to the five major foreign opinion websites, which may result in limited representativeness and diversity of the dataset. In addition, since the study focuses on user-posted content on social media platforms, it may fail to fully capture all relevant perceptual and cognitive dimensions. This limitation of data sources may affect the broad applicability and in-depth analysis of the study.

For the direction of future research, a wider and diverse range of data sources could be considered as a first step. For example, in addition to social media data, more multifaceted data such as on-site interviews, questionnaire surveys, and expert evaluations can be included to improve the comprehensiveness and precision of the study. In addition, future studies can also consider conducting longitudinal studies on the aesthetic changes of specific gardens in different years in order to explore the aesthetic and cognitive evolution in the time dimension.

Second, future studies could also explore in greater depth the effects of single elements (e.g., water features, vegetation, architecture, etc.) on perception and cognition. Through this analysis, researchers can better understand the role of individual design elements in shaping spatial experiences. This not only helps to deepen the understanding of perceptual and cognitive interactions, but also provides more specific guidance for optimizing spatial design.

Finally, given the potential of deep learning techniques in multimodal data analysis, future research could further explore the application of these advanced techniques in the field of spatial design and cultural studies, especially in terms of efficiency and accuracy when dealing with large-scale datasets. This will not only extend the research methodology, but also bring new perspectives and innovative ideas to the field of architecture and landscape design.

## 6 Conclusion

This study empirically analyzes the audience perception of overseas Chinese gardens, revealing a complex system containing multiple scenarios of cognitive responses and multi-level information dimensions. Through the multimodal analysis of social media data, this paper depicts the general trend of overseas Chinese garden audience perceptions. It demonstrates the feasibility of exploring cognitive patterns in specific scenarios. The core contribution of this study is to propose a new approach to understanding and analyzing audience perceptions in landscape gardening, which provides valuable insights into the practice of garden design and cultural communication.

In addition, incorporating AI agents and a Multi-Agent System (MAS) provides new dimensions and perspectives for studying aesthetic cognition. The incorporation of AI agents enhances the depth and breadth of the data analysis, while the application of MAS provides a more dynamic and interactive environment for the study, which makes it possible to understand and cope with the complexity of audience cognition. This provides a theoretical framework for understanding audience perceptions in cross-cultural environments and practical guidance for landscape designers and cultural communicators in practice.

In summary, the findings of this paper not only provide new perspectives and methods for the study of audience perceptions in the field of landscape architecture but also provide valuable guidance for future landscape design and cultural communication practices. Combining different research questions and methods can also apply this research framework to a broader range of research areas, providing new tools for understanding and addressing complex audience perception challenges.